\documentclass{article}
\usepackage{spconf,amsmath,graphicx,hyperref}
\usepackage{comment}
\usepackage{placeins}
\usepackage{multirow}
\usepackage{xurl}

\newif\ifdraft
\drafttrue  
\ifdraft

\newcommand{\deleted}[1]{\textcolor{red}{\stkout{#1}}}

\newcommand{\deletedfloat}[1]{}
\newcommand{\commented}[1]{\textcolor{blue}{#1}}
\else

\newcommand{\deleted}[1]{}

\newcommand{\deletedfloat}[1]{}
\newcommand{\commented}[1]{}
\fi

\title{Explainable Diabetic Retinopathy Classification Using Vision Foundation Models}
\name{Abhishek Verma, Anila Krishna, Abhishek Gajanan Bankar, Juan Miguel Lopez Alcaraz}
\address{Department of Health Services Research, Carl von Ossietzky Universität Oldenburg\\
Oldenburg, Germany \\
Corresponding author: \texttt{juan.lopez.alcaraz@uol.de}}

\begin{document}
%
\maketitle

\begin{abstract}
Diabetic retinopathy (DR) is a leading cause of preventable blindness, and automated screening must be both accurate and trustworthy. We study an explainable DR classification framework built on vision foundation models and three transfer learning strategies. Three backbones, DINOv2, CLIP, and Vision Transformer (ViT), were adapted by full fine-tuning, linear probing, and Low-Rank Adaptation (LoRA). Models were trained and internally evaluated on the ODIR dataset and externally evaluated on APTOS to assess generalization. DINOv2-LoRA achieved the highest internal AUROC (0.758), whereas DINOv2 and ViT full fine-tuning achieved the highest external AUROC (0.920). Probability calibration was assessed with reliability analysis after isotonic regression. For explainability, Grad-CAM and HiResCAM maps were compared with expert-annotated lesion masks from the IDRiD dataset using Dice, Intersection over Union (IoU), and Pointing Game metrics. The results show that foundation models, particularly DINOv2, provide strong predictive performance, and that LoRA is a parameter-efficient alternative to full fine-tuning. Spatial overlap between explanation maps and annotated lesions was limited (mean Dice of about 0.13 for the combined lesion mask), which highlights the value of quantitatively testing whether model attention corresponds to clinically relevant retinal lesions.
\end{abstract}


\section{Introduction}
\label{sec:intro}

\subsection{Clinical importance of diabetic retinopathy diagnosis}

Diabetic retinopathy (DR) is a major complication of diabetes mellitus and one of the leading causes of vision impairment and blindness worldwide \cite{wong2016}. Early detection is essential to prevent irreversible retinal damage and to allow timely treatment \cite{schreur2022imaging}. Fundus photography is widely used for DR screening because it is a non-invasive and cost-effective modality that captures retinal lesions such as microaneurysms, hemorrhages, and exudates \cite{kanclerz2021imaging}.

Manual grading of fundus images, however, requires trained ophthalmologists and is time-consuming, which makes large-scale screening programs difficult to run efficiently. Artificial intelligence (AI) has therefore emerged as a promising route to fast, consistent, and scalable DR detection. Deep learning systems have reached expert-level performance in large screening settings \cite{gulshan2016,ting2017}, and an autonomous AI system has been evaluated prospectively in primary care offices \cite{abramoff2018pivotal}. The rapidly growing literature on fundus-based DR detection and grading has been surveyed in several recent reviews \cite{ikram2024systematic,imran2023icosst}.

\subsection{Clinical foundation models}

Foundation models learn transferable representations from large-scale data and can then be applied to many downstream tasks, which has made them an important paradigm in artificial intelligence \cite{bommasani2021}. They have shown strong potential across healthcare, including medical imaging and physiological signal analysis, where they facilitate transfer to downstream clinical tasks \cite{zhou2024,foundationalclinicalmodels}.

In ophthalmology, foundation models pretrained on retinal images have shown great promise for disease detection and risk prediction \cite{retfound2023}. Vision foundation models more generally have become popular because contrastive and self-supervised pretraining yield highly transferable representations \cite{radford2021,oquab2023}. This ability to generalize makes them attractive for medical image classification, including the classification of diabetic retinopathy.

Pretrained models can be adapted to a downstream task with different transfer learning strategies, which are widely used in DR classification to cope with limited annotated data \cite{raghu2019transfusion,bilal2022transfer}. Full fine-tuning updates all model parameters and often gives strong task-specific performance. Parameter-efficient approaches such as Low-Rank Adaptation (LoRA) \cite{hu2022} instead update only a small subset of parameters; they reduce the number of trainable parameters and can lower the optimization burden while remaining competitive in predictive performance.

\subsection{Explainable AI for trustworthy clinical decision support}

Foundation models can classify diabetic retinopathy with high accuracy, yet their decision process often remains opaque. In clinical practice, clinicians must understand and trust the rationale behind a prediction before an AI system can be integrated into a diagnostic workflow. Explainability has consequently become a key requirement for trustworthy, clinically deployable medical AI \cite{gunning2019,vandervelden2022}.

Explainable AI (XAI) techniques aim to make models more transparent by showing which image regions and features influence a prediction \cite{tjoa2020,henna2025interpretable}. Among the most widely used approaches are gradient-based and activation-based visualization methods, which generate heatmaps of the regions that contribute most strongly to a decision. Grad-CAM is one of the most popular of these methods because it is flexible across architectures and gives intuitive visual explanations \cite{selvaraju2017}.

In DR classification, explanation maps should highlight clinically significant lesions such as microaneurysms, hemorrhages, and exudates. If explanations agree with existing pathological knowledge, they can support validation and error analysis and increase confidence in AI-enabled screening. They must nevertheless be interpreted with caution, because visual saliency is not the same as causal importance \cite{adebayo2018sanity,holzinger2022causal,pearl2019bookofwhy}.

\subsection{Challenges in evaluating explainability}

Despite the widespread use of XAI in medical imaging, evaluating the quality and reliability of explanations remains an open challenge. Classification performance can be measured with well-established quantitative metrics, but there is no universally accepted standard for deciding whether an explanation is clinically meaningful or faithful to model behavior \cite{samek2021}. Recent work has used quantitative metrics to assess explanation quality systematically across model types and data modalities \cite{xaiquantitative}.

Most existing studies still rely on qualitative inspection by clinicians or domain experts. Expert review gives valuable insight, but it is subjective and varies between observers, which makes it hard to compare explainability methods. Recent clinically oriented studies have therefore emphasized externally validated explainable models, to test whether predictions and their explanations remain meaningful beyond the development dataset \cite{externalneoplasmsgeneralization,alcaraz2025liver,alcaraz2025neurocognitive}. In addition, saliency-based explanations can produce visually plausible heatmaps that do not reflect the true reasoning of the underlying model \cite{draelos2021}.

A more rigorous approach is to compare explanation maps with expert-annotated lesion segmentation masks. For diabetic retinopathy, datasets such as IDRiD provide detailed annotations of microaneurysms, hemorrhages, and exudates, which enables an objective evaluation of explanation localization \cite{porwal2020}.

Such quantitative validation shows whether the highlighted regions correspond to clinically relevant pathology. Localization agreement alone, however, does not establish causal faithfulness: saliency-based methods, including popular ones such as Grad-CAM, offer only a partial view of model reasoning, and their output should be read as evidence of spatial correspondence rather than as a complete account of how the model decides.

\subsection{Contributions}

This work investigates explainable diabetic retinopathy classification with vision foundation models and quantitatively evaluates the quality of the resulting explanations. Our main contributions are:

\begin{itemize}

\item We systematically compare three vision backbones, ViT, CLIP, and DINOv2, under full fine-tuning, linear probing, and Low-Rank Adaptation (LoRA) for binary diabetic retinopathy classification.

\item We evaluate model performance on the ODIR dataset and perform external validation on the independent APTOS dataset to assess generalization across datasets.

\item We compare Grad-CAM and HiResCAM and quantitatively evaluate their explanation maps against expert-annotated lesion masks from the IDRiD dataset using Dice, Intersection over Union (IoU), and Pointing Game accuracy.

\item We investigate the trade-off between predictive performance, parameter-efficient adaptation, calibration, and explanation localization to assess the suitability of foundation models for trustworthy diabetic retinopathy screening.

\end{itemize}

We do not propose a new architecture or a new explainability technique. Instead, we assemble established, state-of-the-art components, namely pretrained vision backbones, three adaptation strategies, external validation, probability calibration, and lesion-based explanation assessment, into one integrated evaluation framework for diabetic retinopathy classification. The framework systematically compares supervised, self-supervised, and vision-language pretrained backbones under full fine-tuning, linear probing, and LoRA. Beyond predictive performance, it evaluates probability calibration on internal and external test sets and quantitatively assesses explanation maps against expert-annotated retinal lesion masks. This combined evaluation shows how discrimination, parameter efficiency, probability reliability, and explanation-to-lesion correspondence trade off against each other.

\begin{figure*}[t]
    \centering
    \includegraphics[width=\textwidth]{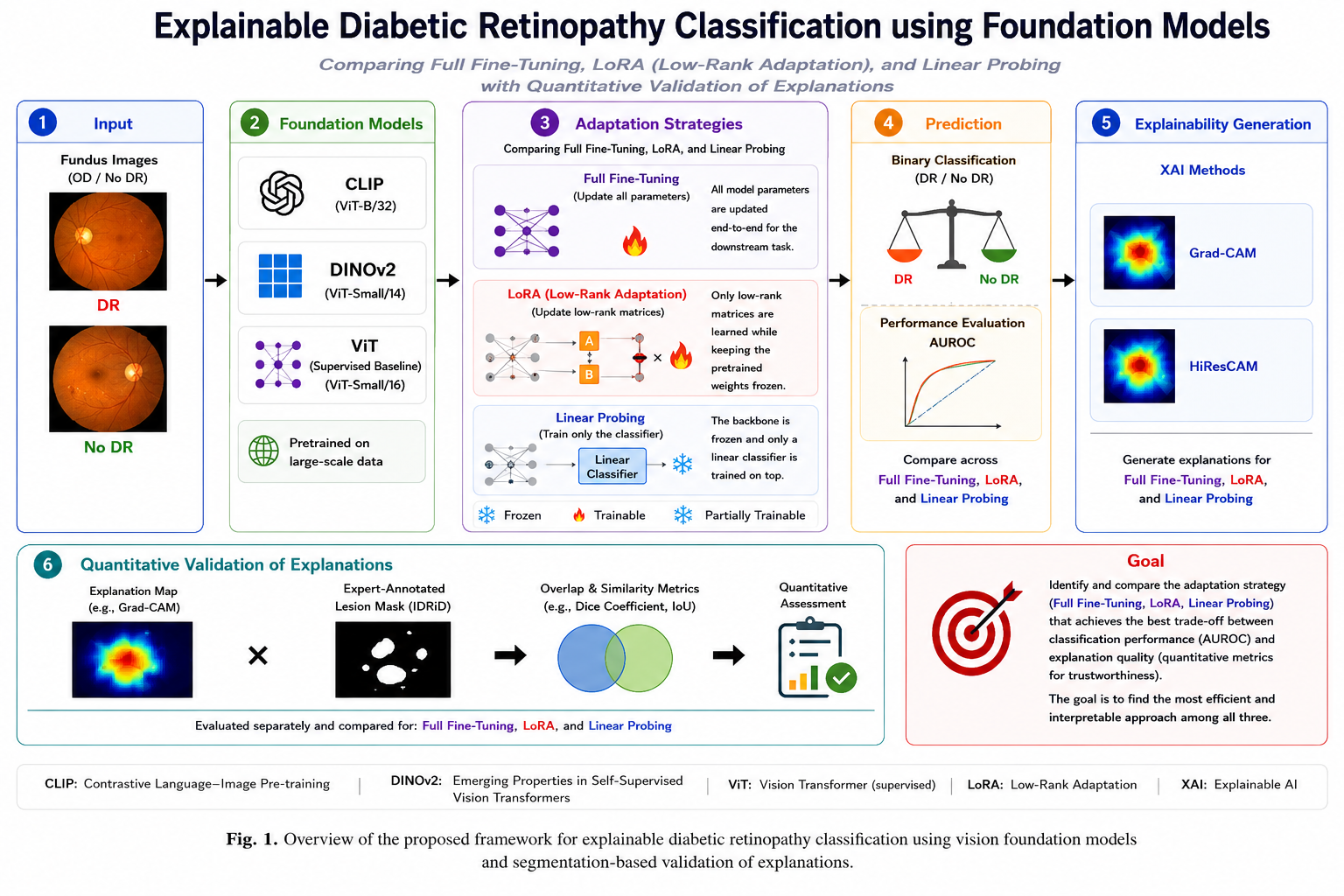}
    \caption{Overview of the proposed framework for explainable diabetic retinopathy classification using vision foundation models and segmentation-based validation of explanations.}
    \label{fig:abstract}
\end{figure*}

\section{Related Work}
\label{sec:related}

\subsection{Deep learning for diabetic retinopathy classification}

Deep learning has substantially advanced automated diabetic retinopathy (DR) detection from retinal fundus images. Early studies showed that convolutional neural networks (CNNs) can learn discriminative retinal features directly from image data and reach performance comparable to expert ophthalmologists in large-scale screening \cite{gulshan2016}. Later work validated these systems across diverse populations and imaging conditions, which supported their use in large-scale screening programs and in improving access to early diagnosis \cite{ting2017}, and demonstrated that autonomous diagnostic systems for DR can be deployed in clinical practice \cite{abramoff2018pivotal}. Transfer learning with pretrained networks has become a common strategy for DR detection from fundus images; one example combines U-Net-based optic disc and blood vessel segmentation with a transfer-learned VGGNet classifier \cite{bilal2022transfer}. Recent reviews cover the datasets, preprocessing, feature extraction, and classification pipelines used for fundus-based DR detection and grading \cite{ikram2024systematic,imran2023icosst}, and new deep learning approaches for detecting and grading DR continue to appear \cite{deepsvdnet2023,ikram2025hybrid}.

More recently, transformer-based architectures have emerged as an alternative to conventional CNNs. Vision Transformers (ViTs) use self-attention to capture long-range relationships within images and perform competitively across many image classification tasks \cite{dosovitskiy2021}.

Despite these advances, most studies focus on maximizing classification accuracy and offer limited insight into interpretability and clinical trustworthiness. Understanding how a model arrives at its prediction therefore remains an important challenge for real-world deployment in healthcare.

Retinal-specific foundation models further demonstrate the value of domain-specific pretraining. RETFound, a self-supervised retinal foundation model pretrained on approximately 1.6 million unlabeled retinal images with a masked-autoencoder objective \cite{he2022masked}, performed strongly across several ocular and systemic disease prediction tasks, including diabetic retinopathy. Its results suggest that representations learned directly from retinal images can improve label efficiency and generalization compared with conventional natural-image pretraining \cite{retfound2023}.

\subsection{Vision foundation models in medical imaging}

Large-scale vision foundation models have recently become powerful representation learners for medical image analysis. Pretraining on large and diverse image collections yields transferable visual features that can be adapted to many downstream clinical tasks, often with substantially fewer labeled samples than conventional supervised approaches \cite{zhou2024}.

Among the most influential vision foundation models are CLIP and DINOv2, which learn visual representations through contrastive learning and self-supervised learning, respectively. Their ability to generalize across domains has led to growing interest in applying them to medical imaging, including disease classification, image segmentation, and clinical decision support \cite{radford2021,oquab2023}.

This success is particularly relevant in healthcare, where collecting and annotating large datasets is expensive, time-consuming, and often restricted by privacy considerations. Retinal foundation models trained on large ophthalmic image collections have shown strong performance across several disease prediction tasks, as well as improved robustness when transferred to new datasets and clinical settings \cite{retfound2023}.

To adapt foundation models efficiently to specialized medical tasks, parameter-efficient fine-tuning approaches such as Low-Rank Adaptation (LoRA) \cite{hu2022} and adapter modules \cite{houlsby2019adapters} have gained attention. They update only a small subset of the model parameters, which greatly reduces computational requirements while preserving much of the performance of full fine-tuning.

\subsection{Explainable AI for diabetic retinopathy analysis}

Explainable AI (XAI) has become an increasingly important part of AI-supported DR diagnosis because it improves model transparency and clinical trust. As models grow more complex, it becomes necessary to understand the reasons behind their decisions, to check that they coincide with clinically important retinal features, and to present them in a form clinicians can readily understand \cite{gunning2019}.

Many approaches have been proposed to visualize the image regions that most influence a prediction. The most common are gradient-based methods such as Grad-CAM, which produce class-specific activation maps that highlight the important features \cite{selvaraju2017}. Newer approaches include HiResCAM \cite{hirescam2021}, Grad-CAM++ \cite{gradcampp2018}, Score-CAM \cite{scorecam2020}, Eigen-CAM \cite{eigencam2021}, and transformer attention visualization \cite{abnar2020}. Explainable AI-driven approaches to automated DR grading have also been proposed \cite{ikram2025explainablegrading}.

For DR screening, explanation maps are expected to agree with the actual retinal pathology, such as microaneurysms, hemorrhages, and exudates. When they do, confidence in the system can increase, errors can be identified, and AI-based screening tools can be verified. It remains controversial, however, whether saliency maps reliably reflect how a model works, since they can look clinically plausible and still be wrong \cite{draelos2021,adebayo2018sanity}.

Beyond Grad-CAM-based methods, several other explanation approaches exist for deep neural networks, including Integrated Gradients \cite{sundararajan2017integrated}, Layer-wise Relevance Propagation (LRP) \cite{bach2015lrp}, and gradient-based saliency maps \cite{simonyan2014saliency}. They differ in how they attribute a prediction to input features and can offer complementary perspectives on model behavior. Their interpretation and quantitative validation remain difficult, however, particularly for medical images in which clinically meaningful regions may be small and spatially heterogeneous. We therefore focus on Grad-CAM and HiResCAM and evaluate their spatial correspondence with expert-annotated retinal lesions, rather than relying on qualitative visualization alone.

\subsection{Quantitative evaluation of explainability}

Visual inspection gives only a qualitative impression of explanation quality. Quantitative assessment is needed to evaluate objectively the robustness and medical relevance of XAI methods. For medical images, it tests whether the explanation map correctly selects the pathological areas that contribute to the model's prediction.

A common strategy is to compare explanation maps with expert-annotated ground-truth lesion masks. Datasets such as IDRiD provide detailed pixel-level annotations of clinically significant lesions, including microaneurysms, hemorrhages, and exudates, so the localization ability of an explanation method can be quantified \cite{porwal2020}. Metrics such as the Dice coefficient (DSC), IoU, and Pointing Game accuracy \cite{zhang2018pointing} measure the similarity between explanation maps and ground-truth annotations, and thus indicate how well methods such as Grad-CAM identify diagnostically relevant retinal structures.

These metrics are useful for quantifying localization, but they do not capture the fidelity of an explanation to the model's underlying decision process. High localization performance does not necessarily mean that the identified regions were causally involved in the prediction \cite{draelos2021}.

\subsection{Research gap}

Previous studies demonstrate the potential of foundation models for retinal disease classification, and explainability methods provide visual insight into model predictions. These two aspects are, however, often evaluated separately, and comparisons across adaptation strategies are limited. In particular, it is not yet clear whether parameter-efficient adaptation such as LoRA can maintain predictive performance while reducing the number of trainable parameters, and whether the resulting explanations correspond to clinically relevant lesions. Evaluation on an independent external dataset is also important to assess generalization beyond the development distribution, and to determine whether both predictive performance and model explanations remain robust across datasets and clinical settings \cite{externalneoplasmsgeneralization,alcaraz2025liver}. Our study addresses these aspects by comparing ViT, CLIP, and DINOv2 across several adaptation strategies, performing external validation on APTOS, and quantitatively evaluating Grad-CAM and HiResCAM against expert-annotated lesion masks from IDRiD.

\section{Methods}
\label{sec:pagestyle}

\subsection{Datasets}

Three publicly available retinal image datasets were used for distinct purposes. ODIR-5K \cite{odir2019} was used for model development and internal evaluation, APTOS 2019 \cite{aptos2019} was reserved for external testing, and IDRiD \cite{porwal2020} was used exclusively for explainability evaluation. Table~\ref{tab:dataset_summary} summarizes the datasets and subsets.

\subsubsection{ODIR-5K}

The ODIR metadata file contained 6,392 rows. After checking the availability of the left- and right-eye fundus images, 12,784 eye-image records were retained, corresponding to 3,358 unique IDs. The images were divided into training (8,946), validation (1,918), and internal testing (1,920) subsets using a 70:15:15 split based on unique IDs and a fixed random seed. Images associated with the same ID were assigned to the same subset to prevent ID-level data leakage.

For binary classification, ODIR images were labelled with the original diabetic retinopathy indicator (D), where D=0 denotes no DR and D=1 denotes DR. The same patient-level label was assigned to the corresponding left- and right-eye images. This resulted in 8,538 non-DR images (66.8\%) and 4,246 DR-positive images (33.2\%). In addition, 747 DR-positive patient records contained labels for other ocular abnormalities.

\subsubsection{APTOS 2019}

For APTOS 2019 \cite{aptos2019}, grade 0 was assigned to the non-DR class and grades 1-4 were grouped into the DR-positive class. All 366 available APTOS images were retained for external testing and were not used for model training or validation. The external evaluation subset comprised 199 non-DR images (54.4\%) and 167 DR-positive images (45.6\%). The original grade distribution was 199 grade-0, 30 grade-1, 87 grade-2, 17 grade-3, and 33 grade-4 images.

\subsubsection{IDRiD}

IDRiD \cite{porwal2020} was used separately because it provides expert-annotated retinal lesion segmentation masks, which are required for the explainability evaluation.

\subsubsection{Image preprocessing}

All images were converted to RGB and resized to $224 \times 224$ pixels. During training, random horizontal and vertical flipping, rotation, and colour jittering were applied to improve robustness and reduce overfitting. Validation and test images were only resized and normalized, using the input statistics of the corresponding pretrained backbone.

\begin{table*}[t]
\centering
\caption{Summary of the retinal image datasets and subsets used in this study. Values in parentheses indicate the percentage of each class within the corresponding subset. ODIR was used for model development and internal evaluation, APTOS for external evaluation, and the IDRiD segmentation subset for explainability evaluation. Quantitative explainability analysis was performed on 49 correctly classified DR-positive images from the 54-image segmentation subset.}
\label{tab:dataset_summary}
\renewcommand{\arraystretch}{1.2}
\begin{tabular}{lccc l}
\hline
\textbf{Dataset} & \textbf{No DR} & \textbf{DR} & \textbf{Total} & \textbf{Role} \\
\hline
ODIR  & 8538 (66.8\%) & 4246 (33.2\%) & 12784 & Development/Internal Test \\
APTOS & 199 (54.4\%) & 167 (45.6\%) & 366 & External Test \\
IDRiD (grading) & 134 (32.4\%) & 279 (67.6\%) & 413 & Grade-label source \\ IDRiD (segmentation) & 0 (0\%) & 54 (100\%) & 54 & XAI Evaluation \\
\hline
\end{tabular}
\end{table*}

\subsection{Models}

Three pretrained vision backbones representing distinct learning paradigms were evaluated. The supervised ViT-Small/16 \cite{dosovitskiy2021} was pretrained on ImageNet and uses a patch size of 16. DINOv2 ViT-Small/14 \cite{oquab2023} uses self-supervised visual representation learning with a patch size of 14. CLIP ViT-Base/32 \cite{radford2021} uses vision-language contrastive pretraining to align image and text representations, with a patch size of 32. The backbones differ in architecture scale and pretraining objective, which allows supervised, self-supervised, and vision-language representations to be compared for diabetic retinopathy classification.

Each backbone was adapted with three strategies: full fine-tuning, linear probing, and Low-Rank Adaptation (LoRA). Full fine-tuning updates all backbone parameters and the classification head, whereas linear probing freezes the pretrained backbone and trains only the classification head. LoRA inserts trainable low-rank updates into selected attention projections while keeping the original backbone weights frozen \cite{hu2022}.

The LoRA configuration used a rank of $r=8$, a scaling factor $\alpha=16$, and a dropout rate of 0.05. For DINOv2 and ViT, LoRA adapters were applied to the query-key-value projection (\texttt{attn.qkv}) and the attention output projection (\texttt{attn.proj}). For CLIP, adapters were applied to the query and value projections (\texttt{q\_proj} and \texttt{v\_proj}). The pretrained weights were kept frozen during LoRA adaptation, while the adapter parameters and the classification head were trainable. Total and trainable parameter counts for all configurations are listed in Table~\ref{tab:parameter_counts}.

\begin{table*}[t]
\centering
\small
\setlength{\tabcolsep}{10pt}
\renewcommand{\arraystretch}{1.15}

\caption{Total and trainable parameter counts for the evaluated model configurations.}
\label{tab:parameter_counts}

\begin{tabular}{l l r r}
\hline
Backbone & Strategy & Total parameters & Trainable parameters \\
\hline
DINOv2 & Full fine-tuning & 22.06M & 22.06M \\
DINOv2 & Linear probing & 22.06M & 385 \\
DINOv2 & LoRA & 22.28M & 0.22M \\
CLIP & Full fine-tuning & 87.46M & 87.46M \\
CLIP & Linear probing & 87.46M & 769 \\
CLIP & LoRA & 87.75M & 0.30M \\
ViT & Full fine-tuning & 21.67M & 21.67M \\
ViT & Linear probing & 21.67M & 385 \\
ViT & LoRA & 21.89M & 0.22M \\
\hline
\end{tabular}
\end{table*}

\subsection{Training setup}

All models were trained on images resized to $224 \times 224$ pixels with a batch size of 32 for a maximum of 10 epochs. We used AdamW optimization \cite{loshchilov2019adamw} with a weight decay of 0.01. The learning rates were $1\times10^{-5}$ for full fine-tuning, $1\times10^{-3}$ for linear probing, and $5\times10^{-5}$ for the LoRA backbone parameters, with $2\times10^{-4}$ for the classification head. A cosine annealing scheduler with a minimum learning rate of $1\times10^{-6}$ was applied. Weighted binary cross-entropy was used, with a positive-class weighting factor derived from the training-set class distribution. Gradient norms were clipped at 1.0, and early stopping with a patience of four epochs was applied on the validation AUROC. A fixed random seed of 42 was used across all experiments.

\subsection{Performance evaluation}

Model performance was primarily evaluated with the area under the receiver operating characteristic curve (AUROC). Calibration was assessed in addition, to evaluate the reliability of predicted probabilities for clinical decision support.

For each model, the classification threshold was selected independently on the ODIR validation set to achieve a sensitivity of at least 85\%. The highest threshold satisfying this target was selected and then applied unchanged to both the ODIR internal test set and the APTOS external evaluation set, so the test sets were never used to determine operating thresholds.

For statistical comparison, paired bootstrap resampling \cite{efron1993bootstrap} was performed with 5,000 resamples to estimate 95\% confidence intervals for AUROC and for pairwise AUROC differences. The same resampled observations were used across all models to preserve paired comparisons. Two-sided bootstrap $p$-values were calculated for each pairwise difference and corrected with the Holm procedure \cite{holm1979} for the 36 comparisons within each dataset. The Holm procedure controls the family-wise error rate across many simultaneous comparisons without assuming that they are independent, and is uniformly more powerful than the Bonferroni correction. Adjusted $p$-values below 0.05 were considered statistically significant. These comparisons quantify the uncertainty of test-set performance estimates; they do not represent variability across independent training runs.

\subsubsection{Calibration analysis}

Calibration was evaluated with the Brier score \cite{brier1950} and the Expected Calibration Error (ECE) \cite{naeini2015ece}, where lower values indicate better calibration \cite{guo2017}. Isotonic regression \cite{niculescu2005predicting} was fitted exclusively on model predictions from the ODIR validation set and then applied unchanged to the ODIR internal test set and the APTOS external test set. No test-set labels or predictions were used to fit the calibrator. Calibration performance was assessed before and after calibration, using 5,000 bootstrap resamples to estimate 95\% confidence intervals. Reliability diagrams were also generated to visualize the agreement between predicted probabilities and observed outcome frequencies.

\subsection{Explainability}

\subsubsection{Grad-CAM}

Grad-CAM \cite{selvaraju2017} produces a class-specific heatmap by weighting the feature maps of a late network layer with the gradients of the target class. The heatmap shows how much each image region contributes to the prediction, which makes it useful for visualizing lesions in diabetic retinopathy diagnosis.

\subsubsection{HiResCAM}

HiResCAM \cite{hirescam2021} is a gradient-based explanation technique and a higher-resolution alternative to Grad-CAM. Pixel-wise importance scores are computed as the element-wise product of feature maps and gradients, which allows precise localization of discriminative areas in the input image. We use it to test whether model attention aligns with disease signs. Grad-CAM and HiResCAM were selected because both generate spatial attribution maps that can be compared directly with the expert-annotated retinal lesion masks available in IDRiD. Using both also allows a conventional gradient-based CAM method to be compared with a higher-resolution variant within the same evaluation framework.

\subsubsection{Attention Rollout (preliminary evaluation)}

Attention Rollout \cite{abnar2020} was explored qualitatively during preliminary experiments. It was excluded from the final quantitative comparison because only Grad-CAM and HiResCAM were systematically evaluated with lesion-localization metrics.

\subsubsection{Quantitative explainability evaluation}

Explanation maps were compared quantitatively with expert-annotated lesion segmentation masks from the IDRiD dataset using the Dice coefficient, Intersection over Union (IoU), and Pointing Game accuracy \cite{zhang2018pointing}. Dice and IoU quantify the spatial overlap between thresholded explanation maps and ground-truth lesion masks, whereas Pointing Game accuracy measures whether the location of maximum activation falls within an annotated lesion region.

The quantitative evaluation included 49 correctly classified DR-positive images from the 54-image IDRiD segmentation subset. Images were selected using the ground-truth DR labels and the model's binary classification threshold of 0.5, and only correctly predicted DR-positive images were retained. No additional confidence threshold was applied. The three high-confidence examples in Figure~\ref{fig:explainability} were selected separately for qualitative illustration and were not part of the quantitative evaluation sample.

For transformer-based visualization, Grad-CAM and HiResCAM were applied to the \texttt{norm1} layer of the final DINOv2 transformer block. The class token was removed, and the remaining 256 patch tokens were reshaped into a $16 \times 16$ spatial feature map before the explanation maps were generated. Each heatmap was normalized by its 99th-percentile activation value and clipped to the range [0,1]. For the Dice and IoU calculations, the normalized heatmaps were binarized with a threshold of 0.5, and the ground-truth masks were resized to $224 \times 224$ pixels. The evaluation considered microaneurysms (MA), hemorrhages (HE), hard exudates (EX), and soft exudates (SE) separately, as well as a combined mask containing all four lesion types. Images without an annotation for a given lesion type were excluded from the calculation of that lesion-specific metric. The optic disc was excluded from the quantitative lesion evaluation.

\section{Results}
\label{sec:results}

\subsection{Predictive performance}

Table~\ref{tab:auroc_models} summarizes the predictive performance of all evaluated configurations, and Figure~\ref{fig:roc} shows the corresponding ROC curves on the internal ODIR and external APTOS datasets. DINOv2-LoRA achieved the highest internal AUROC (0.7581), while ViT-Full achieved the highest external AUROC (0.92011), closely followed by DINOv2-Full (0.92008). DINOv2-LoRA reached an external AUROC of 0.9073, which is competitive cross-dataset performance with far fewer trainable parameters than full fine-tuning. The ROC curves agree with these AUROC values: DINOv2-LoRA shows the strongest internal discrimination, and ViT-Full and DINOv2-Full show the strongest external discrimination.

\begin{table*}[t]
\centering
\scriptsize
\setlength{\tabcolsep}{4pt}
\renewcommand{\arraystretch}{1.35}
\begin{tabular}{lcc}
\hline
\textbf{Model} & \textbf{Internal AUROC (95\% CI)} & \textbf{External AUROC (95\% CI)} \\
\hline

\multicolumn{3}{l}{\textbf{DINOv2 (Self-supervised)}} \\
\hline
DINOv2 (Full Fine-tuning) &
0.755 (0.731, 0.778) &
0.920 (0.886, 0.951) \\

DINOv2 (Linear Probe) &
0.702 (0.677, 0.727) &
0.843 (0.800, 0.884) \\

DINOv2 (LoRA) &
0.758 (0.733, 0.782) &
0.907 (0.872, 0.940) \\

\hline
\multicolumn{3}{l}{\textbf{CLIP (Vision-Language Contrastive)}} \\
\hline
CLIP (Full Fine-tuning) &
0.700 (0.675, 0.725) &
0.755 (0.702, 0.806) \\

CLIP (Linear Probe) &
0.687 (0.661, 0.712) &
0.805 (0.758, 0.850) \\

CLIP (LoRA) &
0.701 (0.675, 0.726) &
0.812 (0.765, 0.854) \\

\hline
\multicolumn{3}{l}{\textbf{ViT (Supervised ImageNet Baseline)}} \\
\hline
ViT (Full Fine-tuning) &
0.740 (0.715, 0.764) &
0.920 (0.890, 0.946) \\

ViT (Linear Probe) &
0.692 (0.666, 0.717) &
0.791 (0.744, 0.838) \\

ViT (LoRA) &
0.715 (0.690, 0.740) &
0.732 (0.679, 0.781) \\

\hline

\end{tabular}
\caption{Performance comparison of all evaluated model-adaptation configurations for binary diabetic retinopathy classification. Models are grouped according to their pretraining paradigm: self-supervised (DINOv2), vision-language contrastive (CLIP), and supervised ImageNet pretraining (ViT). Internal evaluation was performed on the ODIR dataset, while external evaluation was performed on the APTOS dataset. Results are reported as AUROC with 95\% bootstrap confidence intervals.}
\label{tab:auroc_models}
\end{table*}

\begin{figure*}[!t]
\centering
\includegraphics[width=\textwidth]{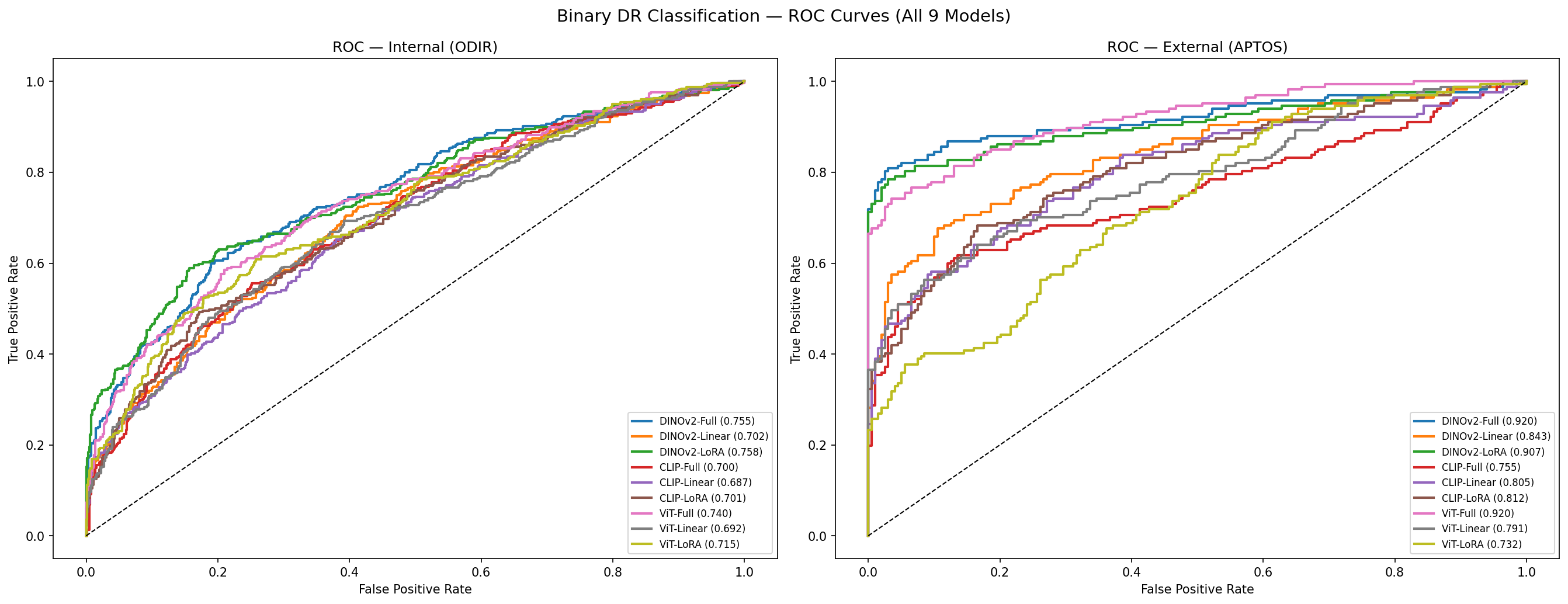}
\caption{ROC curves for the evaluated models on the internal ODIR dataset and the external APTOS dataset.}
\label{}
\end{figure*}

To judge whether AUROC differences are statistically meaningful, we performed paired bootstrap comparisons with 5,000 resamples and applied Holm correction across the 36 pairwise model comparisons within each dataset. After correction, 20 of 36 comparisons were significant on ODIR and 23 of 36 on APTOS. The highest AUROC values were not always significantly different from those of competing models. On ODIR, the AUROC difference between DINOv2-Full and DINOv2-LoRA was small ($-0.0028$) and not significant (95\% CI: $-0.0198$ to $0.0144$; adjusted $p=1.000$). On APTOS, the two models did not differ significantly either (AUROC difference = 0.0128, 95\% CI $-0.0235$ to $0.0494$, adjusted $p=1.000$). Also on APTOS, ViT-Full and DINOv2-Full had nearly identical AUROCs (0.9201 and 0.9201), with no significant difference (95\% CI for the difference: $-0.0247$ to $0.0240$; adjusted $p=1.000$). The observed AUROC ranking should therefore be interpreted together with the uncertainty of the pairwise comparisons.

\subsection{Calibration}

Table~\ref{tab:calibration_metrics} reports the Brier score and ECE before and after isotonic calibration. Isotonic regression reduced ECE for all nine models on both datasets. The Brier score also improved for all models on ODIR and for eight of the nine models on APTOS, with DINOv2-Linear showing a small increase on APTOS (0.180 to 0.190). For DINOv2-LoRA, the Brier score decreased from 0.186 to 0.174 on ODIR and from 0.152 to 0.119 on APTOS, while ECE decreased from 0.114 to 0.072 and from 0.174 to 0.142, respectively. Calibration therefore improved probability reliability overall, although the size of the improvement varied across models and datasets.

\begin{table*}[t]
\centering
\scriptsize
\setlength{\tabcolsep}{4pt}
\caption{Brier score and expected calibration error (ECE) before and after isotonic calibration. Values are means with 95\% bootstrap confidence intervals. Lower values indicate better calibration.}
\label{tab:calibration_metrics}
\begin{tabular}{llcccc}
\hline
Dataset & Model & Brier (Raw) & Brier (Calibrated) & ECE (Raw) & ECE (Calibrated) \\
\hline
\multicolumn{6}{l}{\textit{ODIR internal test set ($N=1,920$)}} \\
& DINOv2-Full   & 0.219 [0.208, 0.230] & 0.175 [0.168, 0.183] & 0.177 [0.165, 0.194] & 0.039 [0.026, 0.056] \\
& DINOv2-Linear & 0.219 [0.211, 0.227] & 0.194 [0.187, 0.200] & 0.152 [0.143, 0.163] & 0.052 [0.038, 0.068] \\
& DINOv2-LoRA   & 0.186 [0.176, 0.196] & 0.174 [0.166, 0.181] & 0.114 [0.098, 0.133] & 0.072 [0.056, 0.090] \\
& CLIP-Full     & 0.265 [0.257, 0.272] & 0.196 [0.189, 0.203] & 0.258 [0.249, 0.267] & 0.049 [0.038, 0.064] \\
& CLIP-Linear   & 0.219 [0.212, 0.226] & 0.198 [0.192, 0.204] & 0.144 [0.136, 0.156] & 0.046 [0.035, 0.063] \\
& CLIP-LoRA     & 0.229 [0.218, 0.240] & 0.194 [0.187, 0.201] & 0.158 [0.144, 0.176] & 0.044 [0.036, 0.064] \\
& ViT-Full      & 0.214 [0.202, 0.226] & 0.181 [0.174, 0.189] & 0.157 [0.141, 0.176] & 0.046 [0.034, 0.065] \\
& ViT-Linear    & 0.216 [0.208, 0.224] & 0.199 [0.193, 0.206] & 0.129 [0.121, 0.144] & 0.071 [0.057, 0.088] \\
& ViT-LoRA      & 0.218 [0.209, 0.228] & 0.188 [0.181, 0.194] & 0.155 [0.143, 0.174] & 0.060 [0.044, 0.078] \\
\hline
\multicolumn{6}{l}{\textit{APTOS external test set ($N=366$)}} \\
& DINOv2-Full   & 0.160 [0.141, 0.179] & 0.108 [0.093, 0.124] & 0.205 [0.182, 0.229] & 0.101 [0.084, 0.134] \\
& DINOv2-Linear & 0.180 [0.168, 0.192] & 0.190 [0.178, 0.202] & 0.164 [0.132, 0.199] & 0.124 [0.101, 0.161] \\
& DINOv2-LoRA   & 0.152 [0.133, 0.171] & 0.119 [0.105, 0.133] & 0.174 [0.155, 0.204] & 0.142 [0.123, 0.172] \\
& CLIP-Full     & 0.246 [0.236, 0.255] & 0.201 [0.188, 0.215] & 0.221 [0.193, 0.256] & 0.102 [0.067, 0.147] \\
& CLIP-Linear   & 0.205 [0.193, 0.216] & 0.195 [0.183, 0.208] & 0.155 [0.136, 0.196] & 0.135 [0.097, 0.178] \\
& CLIP-LoRA     & 0.228 [0.206, 0.251] & 0.177 [0.162, 0.193] & 0.198 [0.177, 0.229] & 0.073 [0.059, 0.108] \\
& ViT-Full      & 0.187 [0.165, 0.210] & 0.126 [0.114, 0.140] & 0.219 [0.194, 0.243] & 0.128 [0.108, 0.151] \\
& ViT-Linear    & 0.201 [0.185, 0.218] & 0.187 [0.172, 0.201] & 0.150 [0.132, 0.187] & 0.093 [0.071, 0.140] \\
& ViT-LoRA      & 0.256 [0.234, 0.278] & 0.210 [0.193, 0.227] & 0.210 [0.190, 0.236] & 0.090 [0.065, 0.137] \\
\hline
\end{tabular}
\end{table*}

\begin{figure}[!b]
\centering
\includegraphics[width=\columnwidth]{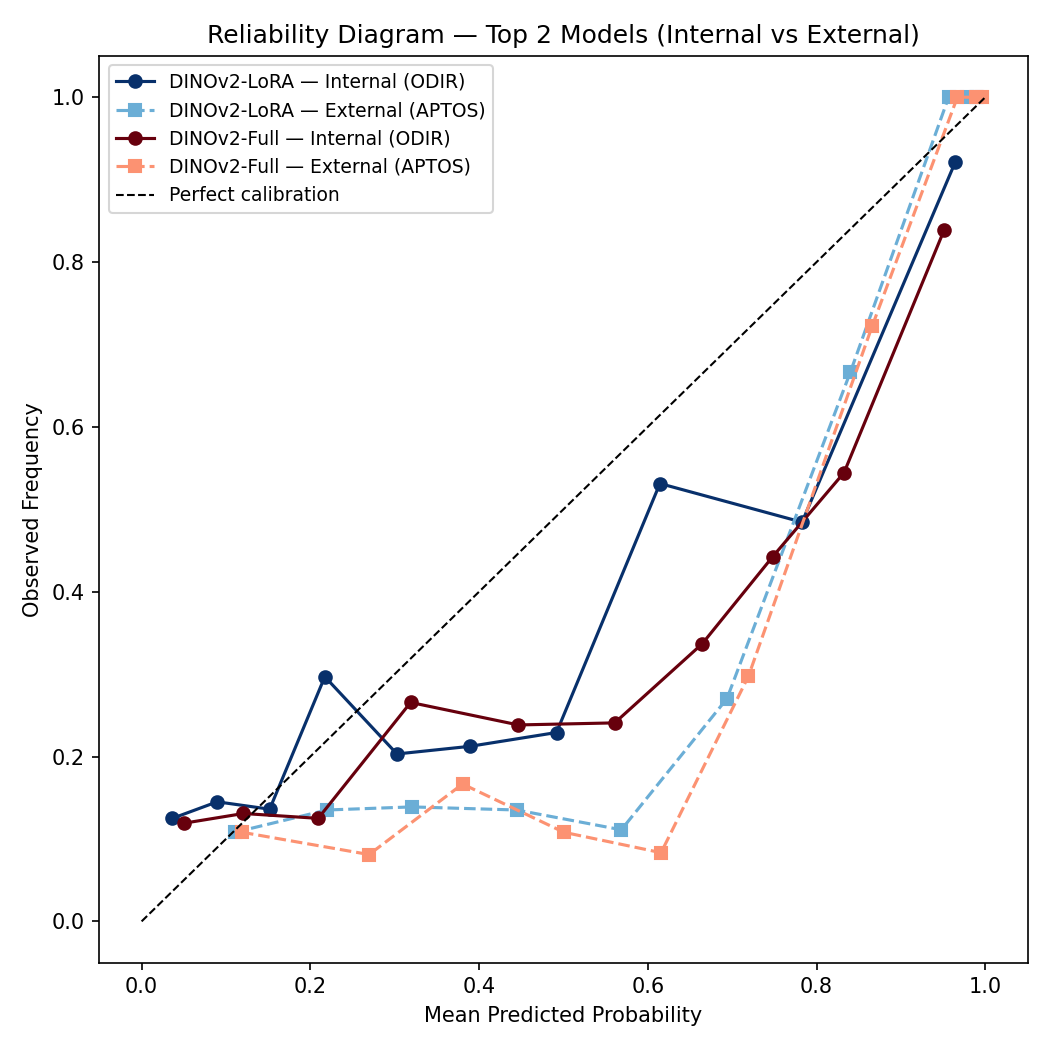}

\caption{Reliability diagrams for the two models with the highest internal AUROC, DINOv2-LoRA and DINOv2-Full, on the ODIR internal and APTOS external datasets, before and after isotonic calibration. The dashed diagonal represents perfect calibration. Isotonic regression was fitted using ODIR validation-set predictions and applied unchanged to the evaluation datasets.}

\label{fig:reliability}
\end{figure}

Figure~\ref{fig:reliability} shows reliability diagrams for DINOv2-LoRA and DINOv2-Full, the two best models by internal AUROC, on the internal ODIR and external APTOS datasets. They visualize the agreement between predicted probabilities and observed outcome frequencies before and after isotonic regression, which was fitted on validation-set predictions.

\subsection{Clinical operating-point performance}

Clinical metrics at the validation-selected thresholds (sensitivity of at least 85\% on the ODIR validation set) are reported in Appendix Table~\ref{tab:clinical_metrics}, and the corresponding confusion matrices in Appendix Table~\ref{tab:confusion_matrices}. On the internal ODIR test set, sensitivity ranged from 79.4\% to 84.8\% and specificity from 35.0\% to 47.8\%. On the external APTOS test set, sensitivity was higher (89.2-99.4\%) but specificity remained low (15.6-33.2\%), which reflects the trade-off between detecting DR-positive cases and avoiding false positives.

\subsection{Comparison with existing approaches}

We compared our results with previously reported retinal foundation-model and transformer-based approaches. RETFound, a retinal-specific foundation model, reported AUROC values of 0.943 on APTOS-2019 and 0.822 on IDRiD for diabetic retinopathy classification \cite{retfound2023}. In the present study, DINOv2-LoRA achieved an AUROC of 0.907 on the external APTOS evaluation, and ViT-Full achieved 0.920. These results are competitive, but direct numerical comparison with published studies should be made cautiously because datasets, preprocessing procedures, class definitions, and evaluation protocols differ across studies. The main contribution of this work is therefore the controlled comparison of supervised, self-supervised, and vision-language foundation models under identical adaptation and external-validation settings.

\subsection{Explainability evaluation}

\begin{figure*}[!t]
    \centering
    \includegraphics[width=\textwidth]{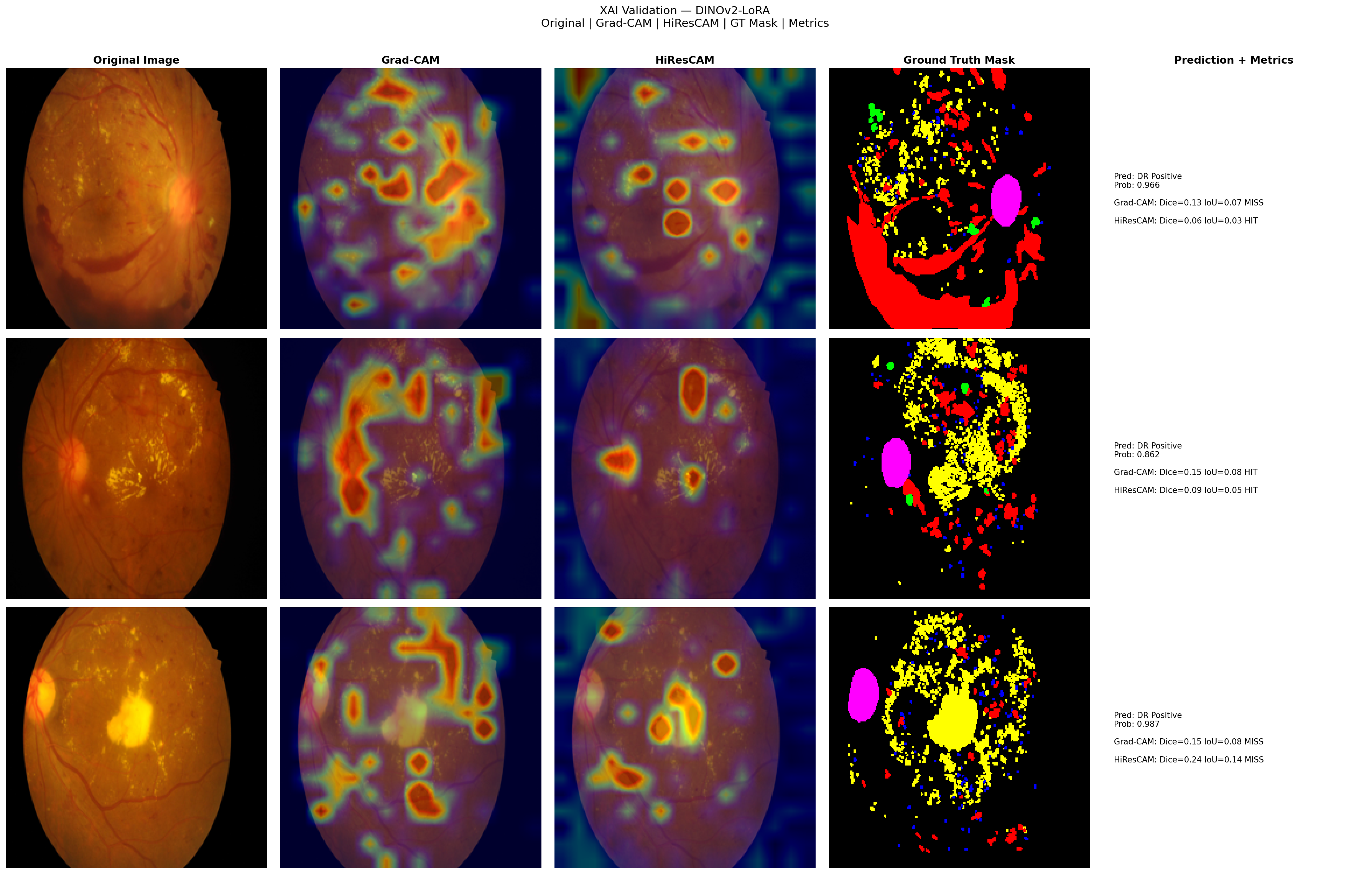}
 \caption{Representative explainability results for the DINOv2-LoRA model on three correctly classified, high-confidence DR-positive IDRiD images (prediction probability above 0.80), selected from the eligible high-confidence cases based on annotated lesion area. From left to right: original fundus image, Grad-CAM, HiResCAM, expert-annotated lesion mask, and model prediction with per-image metrics. The combined lesion mask comprises microaneurysms (MA), hemorrhages (HE), hard exudates (EX), and soft exudates (SE). The optic disc is shown for anatomical reference but excluded from quantitative evaluation. These three images are illustrative examples; cohort-level quantitative results were calculated separately across 49 correctly classified images and are reported in Table~\ref{tab:xai_lesionwise}.}
    \label{fig:explainability}
\end{figure*}

Figure~\ref{fig:explainability} illustrates three qualitative examples. Quantitative explainability evaluation was performed on 49 correctly classified DR-positive IDRiD images. For the combined lesion mask, Grad-CAM achieved a mean Dice of 0.1261 and an IoU of 0.0683, while HiResCAM achieved a mean Dice of 0.1263 and an IoU of 0.0696. The corresponding Pointing Game accuracies were 12.24\% and 16.33\%. Lesion-specific evaluation showed differences across lesion categories: Grad-CAM achieved its highest lesion-specific Dice for hemorrhages (0.0778), whereas HiResCAM achieved its highest for hard exudates (0.0814). The soft exudate evaluation included 26 images, compared with 49 images for each of the other lesion categories. The low Dice and IoU values indicate limited spatial overlap between classification-based saliency maps and expert-annotated lesion regions. Detailed lesion-specific and overall results are given in Table~\ref{tab:xai_lesionwise}.

\begin{table*}[t]
\centering
\small
\setlength{\tabcolsep}{7pt}
\renewcommand{\arraystretch}{1.15}
\caption{Lesion-specific and overall quantitative explainability results on the IDRiD dataset. Dice and IoU are reported as mean values. Pointing Game is reported as the percentage of images in which the maximum-activation location fell within the corresponding lesion mask. N indicates the number of evaluated images.}
\label{tab:xai_lesionwise}
\begin{tabular}{llcccc}
\hline
\textbf{Method} & \textbf{Lesion} & \textbf{N} & \textbf{Dice} & \textbf{IoU} & \textbf{Pointing Game (\%)} \\
\hline
Grad-CAM & MA & 49 & 0.0253 & 0.0129 & 0.00 \\
Grad-CAM & HE & 49 & 0.0778 & 0.0411 & 8.16 \\
Grad-CAM & EX & 49 & 0.0501 & 0.0261 & 2.04 \\
Grad-CAM & SE & 26 & 0.0242 & 0.0124 & 3.85 \\
Grad-CAM & Overall & 49 & 0.1261 & 0.0683 & 12.24 \\
\hline
HiResCAM & MA & 49 & 0.0171 & 0.0087 & 2.04 \\
HiResCAM & HE & 49 & 0.0668 & 0.0356 & 10.20 \\
HiResCAM & EX & 49 & 0.0814 & 0.0445 & 2.04 \\
HiResCAM & SE & 26 & 0.0370 & 0.0201 & 3.85 \\
HiResCAM & Overall & 49 & 0.1263 & 0.0696 & 16.33 \\
\hline
\end{tabular}
\end{table*}

A paired comparison of Grad-CAM and HiResCAM showed no statistically significant difference in overall Dice scores (Wilcoxon signed-rank test, $p=0.622$). The Pointing Game comparison was also not significant (4 Grad-CAM-only hits versus 6 HiResCAM-only hits; exact binomial test, $p=0.754$). The small numerical differences between the two explanation methods should therefore not be read as evidence that one is superior.

\section{Discussion}
\label{sec:discussion}

\subsection{Clinical significance and external generalization}

Our results demonstrate the potential of vision foundation models for automated diabetic retinopathy screening. DINOv2-LoRA achieved the highest internal AUROC, while ViT-Full achieved the highest external AUROC by a very small margin over DINOv2-Full. LoRA provided a parameter-efficient alternative to full fine-tuning with competitive predictive performance. The calibration analysis assessed the reliability of predicted probabilities, which matters in clinical screening workflows where prediction confidence can influence referral decisions \cite{guo2017}. Quantitative evaluation of Grad-CAM and HiResCAM against expert-annotated lesion masks showed only limited spatial correspondence between model explanations and clinically relevant retinal regions. This finding supports using explanations as a complement to, rather than a substitute for, clinical judgment. Because the study combines established components (pretrained vision backbones, parameter-efficient adaptation, calibration, and lesion-based explainability assessment) in a single pipeline, its clinical significance lies in the joint evaluation of discrimination, probability reliability, and explanation quality rather than in a new model. These findings support the potential use of foundation-model-based approaches as decision-support tools for retinal screening, but further clinical validation is required before deployment.

The higher AUROC values on APTOS than on ODIR should be interpreted cautiously. The two datasets differ in sample size, class composition, and other characteristics: APTOS contained 366 external-test images with 45.6\% DR-positive cases, whereas the ODIR internal test set contained 1,920 images with 33.2\% DR-positive cases. Differences in disease severity, image acquisition, image quality, and case selection may also influence discrimination. The underlying patient populations and demographic characteristics may also differ between the datasets; however, these differences were not independently compared in the present study. These factors were not investigated independently in this study, so the higher APTOS AUROC cannot be attributed to any single factor. The smaller external test set also leads to greater uncertainty in the performance estimates. The results therefore show performance on a separate benchmark; they do not establish that external generalization is inherently better than internal performance.

\subsection{Limitations}

This study has several limitations. ODIR and APTOS are public benchmark datasets and may not fully represent real-world clinical variability, and the study addressed binary DR classification rather than severity grading. The analysis was limited to 49 correctly classified IDRiD images, with fewer samples for soft exudates. The low Dice and IoU scores indicate limited spatial overlap rather than accurate lesion segmentation or causal faithfulness. Retinal-specific foundation models were not evaluated. Computational efficiency was assessed only through trainable parameter counts, without measuring training time, memory usage, or inference cost. All experiments used a single random seed and a maximum of 10 epochs, and the bootstrap analysis does not capture training variability across seeds. Future work should address these limitations with broader clinical datasets, retinal-specific models, comprehensive efficiency measurements, and repeated experiments with multiple seeds.

\subsection{Future work}

Future work should evaluate the framework on larger multi-center retinal imaging datasets to assess robustness across patient populations, imaging devices, and acquisition settings. Additional retinal foundation models and multimodal approaches that incorporate clinical information could also be investigated \cite{multimodalclinical,alcaraz2025multimodalbenchmark}. Explainability evaluation could be extended to additional attribution methods and to causal analysis \cite{alcaraz2025causal,holzinger2022causal}, together with more comprehensive lesion-level validation. Finally, extending the framework from binary detection to multiclass disease severity classification, and evaluating it prospectively in clinical settings, would provide further evidence of its clinical utility.

\subsection{Conclusion}

This study evaluated an explainable diabetic retinopathy classification framework based on vision foundation models and three adaptation strategies: full fine-tuning, linear probing, and Low-Rank Adaptation (LoRA). DINOv2-LoRA achieved the highest internal AUROC, while ViT-Full achieved the highest external AUROC by a very small margin over DINOv2-Full. LoRA provided a competitive, parameter-efficient adaptation strategy. Grad-CAM and HiResCAM were further evaluated against expert-annotated lesion masks from IDRiD to assess the spatial correspondence between model explanations and clinically relevant retinal lesions. Overall, the findings demonstrate the potential of foundation-model-based approaches for diabetic retinopathy screening and highlight the importance of evaluating predictive performance, external generalization, calibration, and explainability together. Further multi-center and prospective clinical validation is required before clinical deployment.

\section{Author Contributions}
\label{sec:Author Contributions}

\noindent\textbf{Abhishek Verma:} Software, Methodology, Investigation, Validation, Data Curation, Formal Analysis.

\noindent\textbf{Anila Krishna:} Conceptualization, Methodology, Investigation, Formal Analysis, Visualization, Writing - Original Draft, Writing - Review \& Editing.

\noindent\textbf{Abhishek Gajanan Bankar:} Data Curation, Investigation, Validation, Visualization, Formal Analysis, Writing - Review \& Editing.

\noindent\textbf{Juan Miguel Lopez Alcaraz:} Supervision, Writing - Review \& Editing.

\section{Acknowledgments}
This work is based on coursework in the course "Medical Data Analysis with Deep Learning" carried out under the supervision of Prof. Dr. Nils Strodthoff and Dr. Juan Miguel Lopez Alcaraz.

\section{Code and data availability}

Code Availability: Source code is available at \url{https://github.com/UOLMDA26/retinopathy_vision_foundational}.

Data Availability: The datasets generated during and/or analysed during the current study are available in the ODIR-5K repository \url{https://odir2019.grand-challenge.org/dataset/} \cite{odir2019}, the APTOS 2019 repository \url{https://www.kaggle.com/competitions/aptos2019-blindness-detection/data} \cite{aptos2019}, and the IDRiD repository \url{https://idrid.grand-challenge.org/Data/} \cite{porwal2018idrid}.

\section{Declarations}

\subsection{Ethical approval}
This study involved secondary analysis of publicly available retinal image datasets. The original IDRiD dataset received approval from the local research ethics committee of Shri Guru Gobind Singhji Institute of Engineering and Technology, Nanded, India. The original dataset documentation for ODIR-5K states that patient-identifying information was removed and that its publication followed applicable Chinese ethical and privacy rules. Specific ethics approval details for APTOS 2019 were not available in the dataset documentation, however, it was stated that no patient identifiers, ages, or sexes were included in the public metadata to protect identity. No additional ethical approval was obtained for the present secondary analysis.

\subsection{Consent to participate}
Informed consent was obtained from the participants whose images were included in the original IDRiD dataset. Consent arrangements for ODIR-5K and APTOS 2019 are not specified in the publicly available dataset documentation. No participants were recruited or contacted for the present study.

\subsection{Consent to publish}
Not applicable. This study does not report identifiable individual-level information.

\subsection{Funding}
Not applicable.

\bibliographystyle{IEEEtran}
\bibliography{refs}

\clearpage

\appendix

\section{Additional predictive performance results}
\label{app:additional_results}

\begin{table*}[ht!]
\centering
\scriptsize
\setlength{\tabcolsep}{5pt}
\renewcommand{\arraystretch}{1.15}
\caption{Clinical performance of all nine models at model-specific operating thresholds selected on the ODIR validation set to achieve at least 85\% sensitivity. The selected thresholds were applied unchanged to the ODIR internal and APTOS external test sets. All metrics are reported as percentages.}
\label{tab:clinical_metrics}
\begin{tabular}{lccccccc}
\hline
\textbf{Model} & \textbf{Threshold} & \textbf{Sensitivity/Recall} & \textbf{Specificity} & \textbf{PPV} & \textbf{F1} & \textbf{Accuracy} & \textbf{NPV} \\
\hline
\multicolumn{8}{l}{\textbf{ODIR internal test (N=1,920)}} \\
\hline
DINOv2-Full   & 0.3293 & 84.3 & 46.4 & 43.4 & 57.3 & 58.8 & 85.8 \\
DINOv2-Linear & 0.3517 & 84.8 & 37.2 & 39.7 & 54.1 & 52.8 & 83.3 \\
DINOv2-LoRA   & 0.2423 & 80.5 & 47.8 & 43.0 & 56.0 & 58.5 & 83.4 \\
CLIP-Full     & 0.5105 & 83.7 & 40.5 & 40.7 & 54.8 & 54.6 & 83.5 \\
CLIP-Linear   & 0.3614 & 83.5 & 35.0 & 38.5 & 52.7 & 50.9 & 81.3 \\
CLIP-LoRA     & 0.2547 & 83.5 & 38.1 & 39.7 & 53.8 & 53.0 & 82.6 \\
ViT-Full      & 0.2062 & 84.3 & 40.2 & 40.8 & 54.9 & 54.6 & 84.0 \\
ViT-Linear    & 0.3408 & 79.4 & 39.0 & 38.9 & 52.2 & 52.2 & 79.5 \\
ViT-LoRA      & 0.3024 & 82.4 & 36.3 & 38.7 & 52.7 & 51.4 & 80.8 \\
\hline
\multicolumn{8}{l}{\textbf{APTOS external test (N=366)}} \\
\hline
DINOv2-Full   & 0.3293 & 95.8 & 33.2 & 54.6 & 69.6 & 61.7 & 90.4 \\
DINOv2-Linear & 0.3517 & 94.6 & 32.2 & 53.9 & 68.7 & 60.7 & 87.7 \\
DINOv2-LoRA   & 0.2423 & 95.8 & 27.6 & 52.6 & 67.9 & 58.7 & 88.7 \\
CLIP-Full     & 0.5105 & 89.2 & 20.1 & 48.4 & 62.7 & 51.6 & 69.0 \\
CLIP-Linear   & 0.3614 & 94.0 & 15.6 & 48.3 & 63.8 & 51.4 & 75.6 \\
CLIP-LoRA     & 0.2547 & 95.8 & 18.1 & 49.5 & 65.3 & 53.6 & 83.7 \\
ViT-Full      & 0.2062 & 99.4 & 21.6 & 51.6 & 67.9 & 57.1 & 97.7 \\
ViT-Linear    & 0.3408 & 92.8 & 28.1 & 52.0 & 66.7 & 57.7 & 82.4 \\
ViT-LoRA      & 0.3024 & 97.0 & 16.6 & 49.4 & 65.5 & 53.3 & 86.8 \\
\hline
\end{tabular}
\end{table*}
Table~\ref{tab:clinical_metrics} presents clinical performance at validation-selected operating thresholds. On the internal ODIR test set, sensitivity ranged from 79.4\% to 84.8\%, while specificity ranged from 35.0\% to 47.8\%. On external APTOS evaluation, sensitivity was higher (89.2-99.4\%), but specificity remained lower (15.6-33.2\%). ViT-Full achieved the highest external sensitivity (99.4\%), with one false negative among 167 DR-positive cases. These findings highlight the trade-off between detecting DR-positive cases and avoiding false-positive predictions. The observed test-set sensitivity does not necessarily equal the 85\% validation target because the thresholds were fixed before test evaluation.

\begin{table*}[ht!]
\centering
\caption{Confusion matrices for all models at thresholds selected on the ODIR validation set. The same thresholds were applied to the ODIR internal test set and APTOS external test set.}
\label{tab:confusion_matrices}
\small
\setlength{\tabcolsep}{5pt}
\begin{tabular}{llrrrr}
\hline
Dataset & Model & TN & FP & FN & TP \\
\hline
\multirow{9}{*}{ODIR (N=1920)}
& DINOv2-Full   & 598 & 692 &  99 & 531 \\
& DINOv2-Linear & 480 & 810 &  96 & 534 \\
& DINOv2-LoRA   & 617 & 673 & 123 & 507 \\
& CLIP-Full     & 522 & 768 & 103 & 527 \\
& CLIP-Linear   & 451 & 839 & 104 & 526 \\
& CLIP-LoRA     & 492 & 798 & 104 & 526 \\
& ViT-Full      & 518 & 772 &  99 & 531 \\
& ViT-Linear    & 503 & 787 & 130 & 500 \\
& ViT-LoRA      & 468 & 822 & 111 & 519 \\
\hline
\multirow{9}{*}{APTOS (N=366)}
& DINOv2-Full   & 66 & 133 &  7 & 160 \\
& DINOv2-Linear & 64 & 135 &  9 & 158 \\
& DINOv2-LoRA   & 55 & 144 &  7 & 160 \\
& CLIP-Full     & 40 & 159 & 18 & 149 \\
& CLIP-Linear   & 31 & 168 & 10 & 157 \\
& CLIP-LoRA     & 36 & 163 &  7 & 160 \\
& ViT-Full      & 43 & 156 &  1 & 166 \\
& ViT-Linear    & 56 & 143 & 12 & 155 \\
& ViT-LoRA      & 33 & 166 &  5 & 162 \\
\hline
\end{tabular}
\end{table*}

The confusion matrices provide further insight into the sensitivity-specificity trade-off. On the APTOS external test set, ViT-Full produced the fewest false negatives (1), identifying 166 of 167 DR-positive images, but generated 156 false positives. In contrast, DINOv2-Full produced 7 false negatives and 133 false positives. On the ODIR internal test set, false negatives ranged from 96 to 130 across models, while false positives ranged from 673 to 839. These results highlight that high sensitivity was generally accompanied by a substantial false-positive burden, particularly on the external dataset.

\end{document}